# Monte-Carlo Planning: Theoretically Fast Convergence Meets Practical Efficiency

**Zohar Feldman** and **Carmel Domshlak**
Faculty of Industrial Engineering and Management
Technion, Israel

## Abstract

Popular Monte-Carlo tree search (MCTS) algorithms for online planning, such as $\varepsilon$-greedy tree search and UCT, aim at rapidly identifying a reasonably good action, but provide rather poor worst-case guarantees on performance improvement over time. In contrast, a recently introduced MCTS algorithm BRUE guarantees exponential-rate improvement over time, yet it is not geared towards identifying reasonably good choices right at the go. We take a stand on the individual strengths of these two classes of algorithms, and show how they can be effectively connected. We then rationalize a principle of "selective tree expansion", and suggest a concrete implementation of this principle within MCTS. The resulting algorithms favorably compete with other MCTS algorithms under short planning times, while preserving the attractive convergence properties of BRUE.

## 1 INTRODUCTION

Markov decision processes (MDPs) are a standard model for planning under uncertainty [19]. An MDP $\langle S, A, Tr, R \rangle$ is defined by a set of possible agent states $S$, a set of agent actions $A$, a stochastic transition function $Tr : S \times A \times S \to [0, 1]$, and a reward function $R : S \times A \times S \to \mathbb{R}$. The current state of the agent is fully observable, and the objective of the agent is to act so to maximize its accumulated reward. In the finite horizon setting considered here, the reward is accumulated over some predefined number of steps $H$. The description of the MDP is assumed to be concise, and, depending on the problem domain and the representation language, it can be either declarative or generative (or mixed). While declarative models provide the agents with greater algorithmic flexibility, generative models are more expressive, and both types of models allow for simulated execution of all feasible action sequences, from any state of the MDP.

In online planning for MDPs, the agent focuses on its current state only, deliberates about the set of possible policies from that state onwards and, when interrupted, uses the outcome of that exploratory deliberation to choose what action to perform next. The quality of the action $a$, chosen for state $s$ with $H$ steps-to-go, is assessed in terms of the probability that $a$ is sub-optimal, or in terms of the (closely related) measure of simple regret. The latter captures the performance loss that results from taking $a$ and then following an optimal policy $\pi^*$ for the remaining $H-1$ steps, instead of following $\pi^*$ from the beginning [5].

With a few recent exceptions developed for declarative MDPs [4, 16, 7], most algorithms for online MDP planning constitute variants of what is called Monte-Carlo tree search (MCTS) [23, 18, 15, 9, 8, 21, 25]. Most MCTS algorithms for online planning, such as $\varepsilon$-greedy tree search and UCT, aim at rapidly identifying a reasonably good action, but offer only polynomial-rate reduction of simple regret over the deliberation time. In contrast, a recently introduced MCTS algorithm BRUE guarantees exponential-rate reduction of simple regret over time, yet it does not make special efforts to home in on a reasonable alternative fast [11]. Of course, "good" is often the best one can hope for in large MDPs of interest under practically reasonable deliberation-time allowances. Therefore, as we reconfirm by an evaluation on benchmarks from the recent probabilistic planning competition, BRUE is often empirically inferior to its (guarantees-wise inferior) competitors.

Reflecting on the differences between the two types of algorithms, here we show that BRUE can be redesigned to perform extremely well also under early interruptions of planning, and this without compromising much neither the long-term empirical performance nor theoretical guarantees. We do that in two steps. First, we connect between the iterative tree expansion of the standard MCTS scheme and the "separation of concerns" principle that underlies BRUE. The resulting modification of BRUE, $\text{BRUE}_\mathcal{I}$, already substantially improves over BRUE in short-term effectiveness. Building upon $\text{BRUE}_\mathcal{I}$, we then introduce a machinery of *selective tree expansion* that further pushes the boundaries of online MDP planning. Viewing MCTS as a message passing within the hierarchy of forecasters, this mechanism is based on (i) classifying the roles that different forecasters in the hierarchy should fulfill in order to improve the quality of the decision at the root, and on (ii) exploit this classification to adaptively decide *how* each forecaster should aim at fulfilling its role. As testified by our empirical evaluation, the resulting algorithm, $\text{BRUE}_{\mathcal{IC}}$, favorably and robustly competes with other MCTS algorithms under short planning times, while preserving both the attractive formal properties of BRUE, as well as the empirical strength of the latter under permissive deliberation-time allowances.

## 2 BACKGROUND

Henceforth, $\mathbf{\Pi}$ denotes the set of all valid policies for the MDP in question, $A(s) \subseteq A$ denotes the actions applicable in state $s$, the operation of drawing a sample from a distribution $\mathcal{D}$ over set $\aleph$ is denoted by $\sim \mathcal{D}[\aleph]$, $\mathcal{U}$ denotes uniform distribution, and $[\![n]\!]$ for $n \in \mathbb{N}$ denotes the set $\{1, \ldots, n\}$.

### 2.1 CANONICAL MCTS SCHEME

MCTS, a canonical scheme underlying various MCTS algorithms for online MDP planning, is depicted in Figure 1a. Starting with the current state $s_0$, MCTS performs an iterative construction of a tree[1] $\mathcal{T}$ rooted at $s_0$. At each iteration, MCTS rollouts a state-space sample $\rho$ from $s_0$, which is then used to update $\mathcal{T}$. First, each state/action pair $(s, a)$ is associated with a counter $n(s, a)$ and a value accumulator $\widehat{Q}(s, a)$, both initialized to 0. When a sample $\rho$ is rolled out, for all states $s_i \in \rho \cap \mathcal{T}$, $n(s_i, a_{i+1})$ and $\widehat{Q}(s_i, a_{i+1})$ are updated on the basis of $\rho$ by the UPDATE-NODE procedure. Second, $\mathcal{T}$ can also be expanded with any part of $\rho$; The standard choice is to expand $\mathcal{T}$ with only the first state along $\rho$ that is new to $\mathcal{T}$. In any case, once the sampling is interrupted, MCTS uses the information stored at the tree's root to recommend an action to perform in $s_0$.

---
[1]In MDPs, there is no reason to distinguish between nodes associated with the same state at the same depth. Hence, the graph $\mathcal{T}$ constructed by MCTS instances typically forms a DAG. Nevertheless, for consistency with prior literature, we stay with the term "tree".

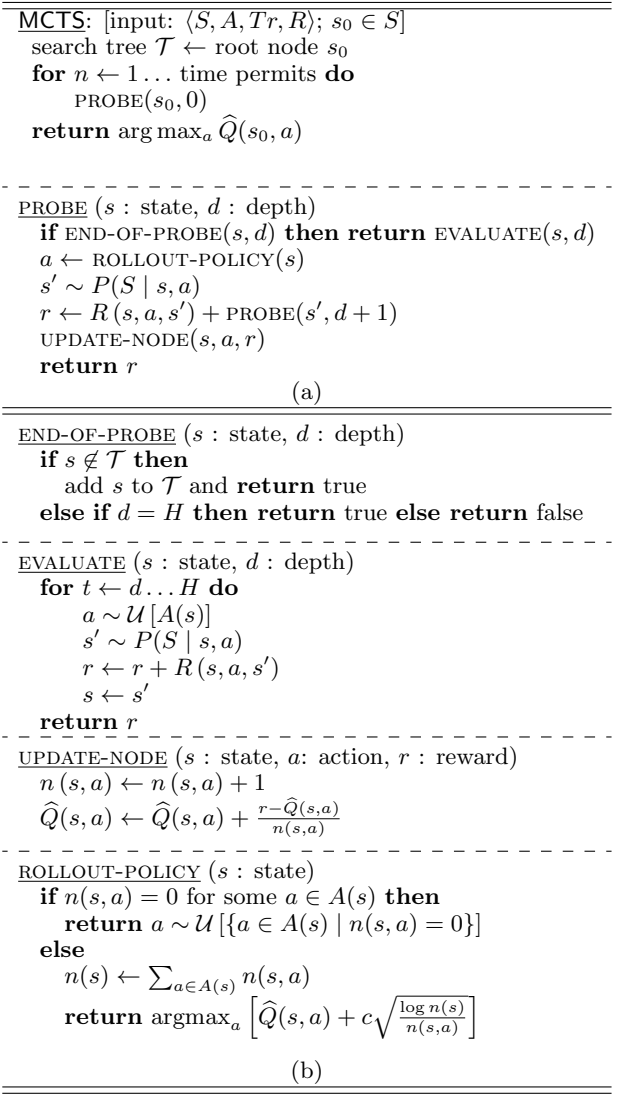

Figure 1: (a) Monte-Carlo tree search template, and (b) the UCT specifics.

Numerous concrete instances of MCTS have been proposed, with $\varepsilon$-greedy [23] probably being the most widely known, and UCT [15] and its modifications [9, 24] being the most popular such instances these days [12, 22, 3, 2, 10, 14]. Concrete instances of MCTS vary mostly along the implementation of the ROLLOUT-POLICY sub-routine, that is, in their policies for directing the rollout within $\mathcal{T}$. For instance, the specific ROLLOUT-POLICY of UCT is shown in Figure 1b. This policy is based on the deterministic decision rule UCB1 [1], originally proposed for optimal balance between exploration and exploitation for cumulative regret minimization in stochastic multi-armed bandit (MAB) problems [20]. In general, different instances of MCTS vary in their balance between exploration and exploitation. However, it has already been noticed that exploitation may considerably slow down the reduction of simple regret over time [6]. In-

```
MCTS2e: [input: ⟨S, A, Tr, R⟩; s₀ ∈ S]
  search tree 𝒯 ← root node s₀; σ ← 0
  for n ← 1 ... time permits do
    σ ← SWITCH-FUNCTION(n, σ)
    PROBE(s₀, 0, σ)
  return arg max_a Q̂(s₀, a)
```
- - - - - - - - - - - - - - - - - - - - - - - - - - -
```
PROBE (s : state, d : depth, σ ∈ ⟦H⟧)
  if END-OF-PROBE(s, d) then return EVALUATE(s, d)
  if d < σ then
    a ← EXPLORATION-POLICY(s)
  else
    a ← ESTIMATION-POLICY(s)
  s' ∼ P(S | s, a)
  r ← R(s, a, s') + PROBE(s', d + 1, σ)
  if d = σ then UPDATE-NODE(s, a, r)
  return r
```
(a)

```
END-OF-PROBE (s : state, d : depth)
  if d = H then return true else return false
```
- - - - - - - - - - - - - - - - - - - - - - - - - - -
```
EVALUATE (s : state, d : depth)
  return 0
```
- - - - - - - - - - - - - - - - - - - - - - - - - - -
```
UPDATE-NODE (s : state, a : action, r : reward)
  if s ∉ 𝒯 then add s to 𝒯
  n(s, a) ← n(s, a) + 1
  Q̂(s, a) ← Q̂(s, a) + (r − Q̂(s, a))/n(s, a)
```
- - - - - - - - - - - - - - - - - - - - - - - - - - -
```
SWITCH-FUNCTION (n : iteration, σ ∈ ⟦H⟧)
  return H − ((n − 1) mod H)   // round robin on ⟦H⟧
```
- - - - - - - - - - - - - - - - - - - - - - - - - - -
```
EXPLORATION-POLICY (s : state)
  return a ∼ 𝒰[A(s)]
```
- - - - - - - - - - - - - - - - - - - - - - - - - - -
```
ESTIMATION-POLICY (s : state)
  return a ∼ 𝒰 [{a | arg max_{a ∈ A(s)} Q̂(s, a)}]
```
(b)

Figure 2: Monte-Carlo tree search with "separation of concerns" (a), and the BRUE specifics (b).

deed, UCB1 (and thus UCT) achieves only polynomial-rate reduction of simple regret over time [6], and the number of samples after which the bounds of UCT on simple regret become meaningful might be as high as hyper-exponential in $H$ [9]. In fact, no instance of the MCTS scheme (Figure 1a) suggested so far breaks the barrier of the worst-case polynomial-rate reduction of simple regret over time.

## 2.2 SEPARATION OF CONCERNS

If fast convergence to optimal choice is of interest, then Monte-Carlo planning should be as exploratory as possible [6]. However, what it means to be "as exploratory as possible" with MDPs is less straightforward than it is in MABs. In particular, recently it was observed that "forecasters" $s \in \mathcal{T}$ should be devoted to *two*, somewhat competing, *exploratory* objectives, namely identifying an optimal action $\pi^*(s)$, and estimating the value of that action, because this information is needed by the predecessor(s) of $s$ in $\mathcal{T}$ [11].

Following this observation, Feldman and Domshlak [11] introduced MCTS2e, a refinement of MCTS scheme that implements the principle of "separation of concerns," whereby different parts of each sample are devoted to different exploration objectives. In MCTS2e (Figure 2a), rollouts are generated by a two-phase process in which the actions are selected according to an exploratory policy until an (iteration-specific) switching point, and from that point on, the actions are selected according to an estimation policy. The sub-routines END-OF-PROBE and UPDATE in MCTS2e are trivial, the UPDATE-NODE sub-routine extends this of MCTS with tree expansion, and, instead of ROLLOUT-POLICY of MCTS, specific instances of MCTS2e are achieved by instantiating three new sub-routines that determine the switching point for a rollout, and the action selection protocols for the two phases of the rollouts.

Feldman and Domshlak [11] show that a specific instance of MCTS2e, dubbed BRUE, achieves an *exponential-rate* reduction of simple regret over time, with the bounds on simple regret becoming meaningful after only exponential in $H^2$ number of samples. The specific MCTS2e sub-routines that define the BRUE algorithm are shown in Figure 2b. Similarly to UCT, each node/action pair $(s, a)$ is associated with variables $n(s, a)$ and $\widehat{Q}(s, a)$, but with the latter being initialized to $-\infty$. BRUE instantiates MCTS2e by choosing actions uniformly at the exploration phase of the sample, choosing the best empirical actions at the estimation phase, and changing the switching point in a round-robin fashion over the entire horizon. Importantly, if the switching point of a rollout $\rho = \langle s_0, a_1, s_1, \ldots, a_H, s_H \rangle$ is $\sigma$, then only the state/action pair $(s_{\sigma-1}, a_\sigma)$ is updated by the information collected by $\rho$. That is, the information obtained by the estimation phase of $\rho$ is used only for improving the estimate at state $s_{\sigma(n)-1}$, and is not pushed further up the sample. While that may appear wasteful and counterintuitive, this locality of update is required to satisfy the formal guarantees of BRUE on exponential-rate reduction of simple regret over time [11].

## 3 FAST OPTIMAL VS. FAST GOOD

A comparative evaluation on *Sailing* [18] and *PGame* [15] domains showed that BRUE is continually improving towards an optimal solution, rather quickly obtaining results better than UCT [11]. However, that evaluation also showed that UCT sometimes manages to identify reasonably good actions rather quickly, while BRUE is still "warming up". Focusing on tight deliberation deadlines, we conducted a wider empirical evaluation on MDP benchmarks from the last International Probabilistic Planning Competition (IPPC-2011). These benchmarks appear ideal for our purpose of evaluating algorithms under tight time constraints: Most IPPC-2011 domains induce very large branching factors, and thus allow only a very shallow sampling of the underlying search tree in reasonable time.

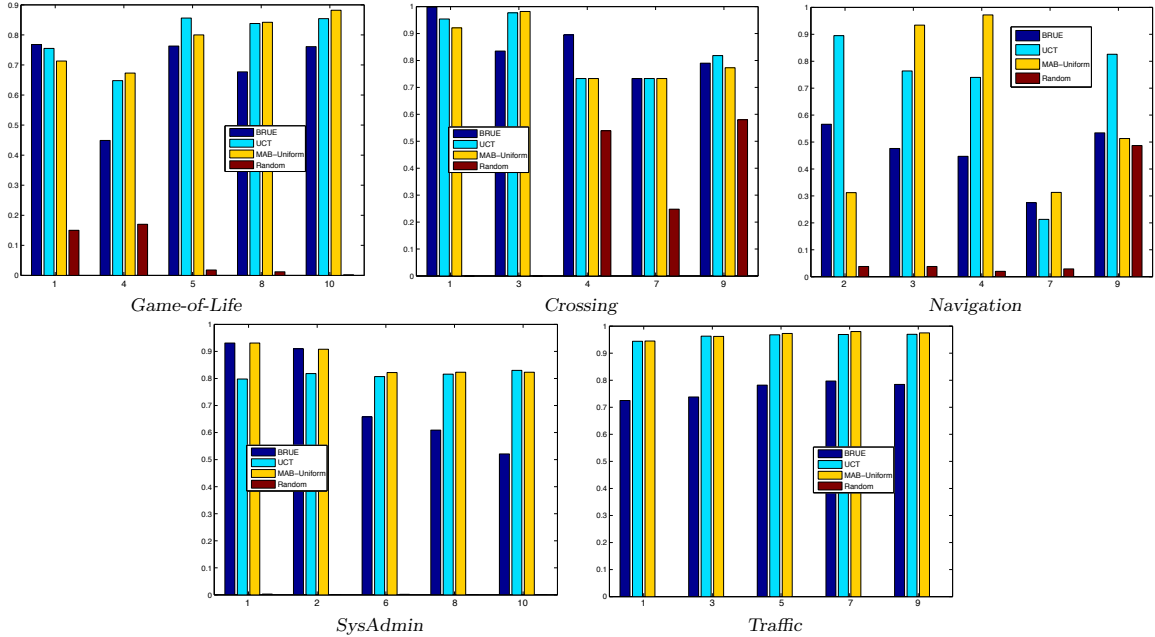

Figure 3: IPPC-2011 scores for the different MCTS algorithms under the (unknown to the algorithms) average deliberation time allowance of 5 seconds per step.

We used five IPPC-2011 domains, *Game-of-Life*, *SysAdmin*, *Traffic*, *Crossing*, and *Navigation*, with five randomly picked problem instances from each domain. Instances of *Game-of-Life*, *SysAdmin*, and *Traffic* were ran with planning horizon of 20 steps, whereas instances of (goal-driven) *Crossing*, and *Navigation* were ran with horizon of 40 steps. Each algorithm was allowed a (rather arbitrary chosen) deliberation time that started with 10 seconds for the first step, and decreased linearly to 1 second at the last step. In addition to UCT and BRUE, the comparison includes a trivial baseline of random action selection (Random), as well as MAB-Uniform, a simple algorithm that chooses actions everywhere uniformly at random, estimating $Q(s_0, a)$ for each $a \in A(s_0)$ by the value of taking $a$ and then choosing actions uniformly at random for the remaining $H-1$ steps. Obviously, the estimates of MAB-Uniform are typically erroneous, and so a priori, its only positive property appears to be its computational efficiency.

The results are presented in Figure 3; the names of the problem instances on the $x$-axis are these from the IPPC-2011 repository, and $y$-axis captures the IPPC-2011 scoring scheme: The algorithms are scored relatively to each other on each problem instance, with the relative scores being then averaged over 300 runs on that specific problem instance. The relative score of a particular algorithm in a specific run is the difference between the total reward achieved by that algorithm and the worst total reward among all the algorithms, divided by the difference between such best and worst total rewards.

According to Figure 3, all the examined MCTS algorithms, including the seemingly naive MAB-Uniform, performed much better than Random, with *SysAdmin* and *Traffic* being the most prominent examples for that. In other words, even under severely limited deliberation time allowance, the value of deliberation in the benchmarks in use was substantial. Likewise, the experiment reconfirmed that UCT is typically more effective than BRUE under short deliberation times. This suggests that the "fast optimal" scheme of BRUE lacks some ingredients that make its MCTS competitors "fast good". Having said that, note that different problems favored different approaches, with MAB-Uniform being surprisingly superior on many of the examined benchmarks. Hence, the quest for a clear recipe for "fast good" remains open, especially if we do not want to neglect striving for optimality, and even more so, if we want that recipe to be robust on a wide palette of MDPs. This is precisely the quest we consider in what comes next.

# 4 TWO TYPES OF FORECASTERS

Consider the state/steps-to-go pairs $(s, h)$ as a hierarchy of forecasters, all acting on behalf of the root forecaster $(s_0, H)$ that aims at minimizing its own simple regret in a stochastic MAB induced by the applicable actions $A(s_0)$. In the setup of online planning, there is a conceptual difference between the exploration objective of the root forecaster and this of all other forecasters in the hierarchy. To see that, suppose there is an oracle that can provide each forecaster $(s, h)$ *either* with the identity of the optimal action $\pi^*(s, h)$ but without revealing its value $Q_h(s, \pi^*(s, h))$, *or* with the value $Q_h(s, \pi^*(s, h))$ but without revealing the identity of $\pi^*(s, h)$. For the root forecaster $(s_0, H)$, the first type of information is all he needs, while the second type of information buys him very little, if anything. In contrast, even if the oracle provides *all* the forecasters $(s, h)$ *but* $(s_0, H)$ with (only) the identities of the respective optimal actions $\pi^*(s, h)$, then the root forecaster $(s_0, H)$ in some sense remains as clueless as it was before, and needs to explore the state space in order to obtain at least some ordinal information about the expected value of the alternative choices $A(s_0)$. However, if the oracle provides a non-root forecaster $(s, h)$ (only) with the best $Q$-value among its alternative choices $A(s)$, then $(s, h)$ can stop working since no further exploration of the sub-hierarchy rooted in $(s, h)$ is needed.

In sum, what matters to the root forecaster is only what to execute, while all other forecaster care only about the value they can provide to their ancestors in the hierarchy, and *not about how this value can actually be acquired.* Of course, the reader may question this classification by arguing that these two objectives are just two sides of the same coin: estimating the value of optimal action assumes aiming at identifying an optimal action and vice versa. To some extent, that is true, but only to some extent. For instance, the very realization that this coin has two sides, and that these two sides are somewhat competing, is precisely what motivates the "separation of concerns" principle behind the MCTS2e scheme. Turns out that this classification of objectives suggests further insights into the dynamics of MCTS algorithms.

In all MCTS algorithms for online MDP planning, each iteration corresponds to examining a chain of forecasters within the overall hierarchy under $(s_0, H)$, with the difference between the algorithm boiling down to two decisions:

(I) which chain of forecasters to examine, and

(II) how to estimate $Q_h(s, \pi^*(s, h))$ for each forecaster $(s, h)$ in the hierarchy.

At first view, choosing the right strategy for (I) seems to be the key to rapid homing in on "good" decisions. The details of various MCTS algorithms suggest that their design was indeed primarily guided by choices for (I), with choices for (II) being implied by the former. Here, however, we suggest that decoupling these two decisions is important, and that the key to the quest of our interest actually lies in decision (II).

A closer look at different Monte-Carlo planning algorithms for MDPs reveals an interesting generalizing perspective on the way they all approach decision (II). Let $V_h^\pi(s)$ be the value of $(s, h)$ under policy $\pi \in \mathbf{\Pi}$, $V_h^*(s) \equiv Q_h(s, \pi^*(s, h))$ be the value of $(s, h)$ under the optimal policy, and let $\widehat{V}_h^\pi(s)$, $\widehat{V}_h^*(s)$ denote empirical estimates of these two quantities, respectively. In all MCTS algorithms, at each point of time, the entire hierarchy of forecasters can be seen as consisting of *two types* of forecasters.

$\mathbf{T_{OUT}}$ forecaster $(s, h)$ (possibly schematically) estimate $V_h^*(s)$ by an estimate of $\mathbb{E}_{\pi \sim \mathcal{U}[\mathbf{\Pi}]} V_h^\pi(s)$, that is, of the expected total reward of a policy sampled from $\mathbf{\Pi}$ uniformly at random.

$\mathbf{T_{IN}}$ forecaster $(s, h)$ distinguishes between its choices $A(s)$, and estimates $V_h^*(s)$ by an estimate of $\max_{a \in A(s)} \sum_{s'} P(s' \mid s, a) \left[ R(s, a, s') + V_{h-1}^*(s') \right]$, where the estimate of $V_{h-1}^*(s')$ is based on the information provided by $s'$ to $s$.

Consider the way in which the specific MCTS algorithms approach decision (II) in terms of this $T_{OUT}/T_{IN}$ partition of the forecasters. In both UCT and BRUE, $T_{IN}$-forecasters correspond to the nodes of $\mathcal{T}$, while all other state/steps-to-go pairs correspond to $T_{OUT}$-forecasters. Note that these $T_{OUT}$-forecasters are very much not virtual. For instance, in UCT they are queried by the EVALUATE sub-routine, and in BRUE they are queried, possibly in interleaving with $T_{IN}$-forecasters, by both EXPLORATION-POLICY and ESTIMATION-POLICY sub-routines.

At first view, $T_{OUT}$-forecasters appear to be strangely lazy and potentially very misleading, while $T_{IN}$-forecasters seem to be doing the right thing. However, it is not all that simple. First, while each $T_{OUT}$-forecaster samples a single random variable, each $T_{IN}$-forecaster $(s, h)$ has to sample $|A(s)|$ random variables. Thus, $T_{OUT}$-forecasters converge to quality estimates of quantities of their interest much faster than their $T_{IN}$ counterparts. Second, while $T_{IN}$-forecasters try to estimate the right thing, their success totally depends on the quality of estimates of $V_{h-1}^*(s')$ they receive from their successors. Hence, it is not clear that forecasters of type $T_{IN}$ are always more effective.

We return to this issue in more detail later on. For now, note only that both UCT and BRUE can be seen as *continuously reconsidering the typing of the forecasters*. Specifically, in both UCT and BRUE, (at most) a single forecaster is "converted" from $T_{OUT}$ to $T_{IN}$ at every iteration: in UCT it is the *shallowest* $T_{OUT}$-forecaster found along the rollout, and in BRUE, it is the $T_{OUT}$-forecaster that happens to lie at the rollout's switching point $\sigma$. This way, the set of $T_{IN}$-forecasters in UCT grows *incrementally* as a single community connected to the root forecaster $(s_0, H)$. In contrast, $T_{IN}$-forecasters in BRUE evolve in $H$ independent sets, each distributed over the respective depth level of the forecast hierarchy according to the transition distribution induced by the *uniform* action selection at the preceding levels.

This specific difference between UCT and BRUE is directly related to their relative efficiency under different orders of deliberation time allowance. Populating $T_{IN}$-forecasters at all levels of the hierarchy is generally necessary to guarantee fast convergence to optimal choice at the root. However, the marginal value of $T_{IN}$-forecasters at different levels vary with the deliberation time allowance: Information gathered by $T_{IN}$-forecasters at deep levels takes time to be propagated to the root, making their near-term influence on the choices at $(s_0, H)$ smaller than this of the $T_{IN}$-forecasters closer to the root.

In that respect, a modification of BRUE that suggests itself almost immediately is as simple as it gets: Instead of converting the $T_{OUT}$-forecaster at the switching point $\sigma$, we can resort to converting the shallowest $T_{OUT}$-forecaster on the exploratory part of the rollout, that is, up to the level $\sigma$. By offering both exponential-rate reduction of the simple regret at the root, as well as incremental conversion of $T_{OUT}$-forecasters as a connected set around $(s_0, H)$, the resulting algorithm, BRUE$_\mathcal{I}$, substantially improves over BRUE in short-term effectiveness. (The specific empirical results for BRUE$_\mathcal{I}$ are shown later in the paper.) However, this simple modification of BRUE is not our final destination, and next we show that this simple bridge between MCTS and MCTS2e opens a much wider window of opportunity.

## 5 SELECTIVE TREE EXPANSION

Similarly to UCT and BRUE, each iteration of BRUE$_\mathcal{I}$ either finds no candidate for type conversion, or *unconditionally* converts a concrete single $T_{OUT}$-forecaster $(s, h)$ to type $T_{IN}$. However, suppose we know that $\mathbb{E}_{\pi \sim \mathcal{U}[\mathbf{\Pi}]} V_h^\pi(s)$ *equals* $\max_{a \in A(s)} \sum_{s'} P(s' \mid s, a) \left[ R(s, a, s') + V_{h-1}^*(s') \right]$. Since direct Monte-Carlo estimation of the quantity on the left-hand side is substantially easier than this of the right-hand side, converting $(s, h)$ to type $T_{IN}$ is clearly not a good idea. In fact, both $(s, h)$ and all of its exclusive descendants in the hierarchy would better remain $T_{OUT}$-forecasters for the entire deliberation process, no matter how long it is. Of course, this equality rarely holds, and, more importantly, we have no prior knowledge about the size of the gap between the quality of the best policy under $(s, h)$ and the expected quality of the randomly picked policy. However, this extreme example still hints on the promise of *selective* type conversion, and below we examine the prospects of this direction.

The variance of a Monte-Carlo estimator $\widehat{Q}_h(s, a)$ of the value of action $a$ at state $s$ stems from two sources. The first source of variance comes from following different policies (aka action selections) along different rollouts. The other source of variance comes from the stochastic nature of the action outcomes. That is, if $r$ is the reward obtained by following policy $\pi$ for $h$ steps starting from state $s$, then

$$\text{Var}[r] = \mathbb{E}[\text{Var}[r \mid \pi]] + \text{Var}[\mathbb{E}[r \mid \pi]]. \quad (1)$$

At one extreme, we have all policies yielding the same expected reward, and thus all the variance comes from the action outcomes. In that case, distinguishing between the policies under $(s, h)$ is not only useless, but also computationally harmful. Thus both $(s, h)$ and its descendants should be left as type $T_{OUT}$, that is, not added to $\mathcal{T}$. At the other extreme, we have all actions being deterministic, but different policies yield very different reward. In that case, it may be valuable to convert $(s, h)$ to $T_{IN}$, increasing the resolution at which the policies under $(s, h)$ are examined. Unsurprisingly, in between these two extremes, the "value of conversion" is less straightforward. In the absence of any information about the stopping time, online planning algorithms should strive to do well under the assumption that termination point is near, while ensuring continuous improvement as more time is allowed. And as long as the precise mixture of these two desiderata remains vague, so remains the precise formulation of the value of conversion.

Having said that, as we do understand the high-level factors that affect the value of conversion, we can try estimating and combining these factors so to reflect the purported value of conversion. Here we propose and evaluate a simple and intuitive rule: *The candidate $T_{OUT}$-forecaster $(s, h)$ should be converted iff the variance of the expected reward over different policies under $(s, h)$ exceeds the average variance of the policies*, that is, iff

$$\text{Var}\left[\mathbb{E}\left[\widehat{V}_h^\pi(s) \mid \pi\right]\right] \;>\; \mathbb{E}\left[\text{Var}\left[\widehat{V}_h^\pi(s) \mid \pi\right]\right], \quad (2)$$

which is equivalent to

$$\text{Var}\left[\mathbb{E}\left[r \mid \pi\right]\right] > \mathbb{E}\left[\frac{\text{Var}\left[r \mid \pi\right]}{n(s,\pi)}\right], \quad (3)$$

where $r$ is the reward obtained by following policy $\pi$ for $h$ steps starting from state $s$, and $n(s,\pi)$ is the number of samples that induce the estimate $\widehat{V}_h^\pi(s)$.

The quantity on the left indicates the distance between the average quality of the policies and the quality of the optimal one, that is

$$\mathbb{E}_{\pi \sim \mathcal{U}[\Pi]} V_h^\pi(s) - V_h^*(s). \quad (4)$$

The quantity on the right indicates the distance between the estimator and the mean of the policies value, where the division by $n(s,\pi)$ captures the effect of averaging on the variance. As the number of samples from each policy grows, estimators $\widehat{V}_h^\pi(s)$ approaches their means $V_h^\pi(s)$, and thus the gap captured by Eq. 4 becomes the dominant factor, in which case converting $(s,h)$ to $T_{IN}$ is estimated as valuable.

Based on this decision rule, we suggest a new instance of MCTS2e, BRUE$_{\mathcal{IC}}$ (standing for BRUE with *incremental* and *selective* type conversion). The respective MCTS2e sub-routines of BRUE$_{\mathcal{IC}}$ are depicted in Figure 4.

- Similarly to BRUE, the actions are selected uniformly at random at the exploration phase of the rollout, and the empirically best actions are selected at the estimation phase.

- Similarly to UCT and BRUE$_{\mathcal{I}}$, the tree is expanded in an incremental fashion, maintaining the set of $T_{IN}$ forecasters connected to the root.

- Unlike UCT and BRUE$_{\mathcal{I}}$, after a forecaster $(s,h)$ is added to the tree, it goes through an "evaluation period", and remains of type $T_{OUT}$ until it passes that evaluation. Hence, the leaves of $\mathcal{T}$ consist of some $T_{IN}$-forecasters, but also of all $T_{OUT}$ candidates for $T_{IN}$.

The evaluation of the "conditional candidates" $(s,h)$ is captured by the CONVERT procedure depicted in Figure 5. This procedure estimates the two sources of variance at $(s,h)$ (lines 1-4, explained below), and only when the variance of the policies' values exceeds the average variance of the policies value in the sense of Eq. 2 (the condition in line 5 fails), $(s,h)$ is converted to a $T_{IN}$-forecaster. At the actual conversion (lines 6-9), the information gathered at the evaluation period is used to initialize the standard node variables $\widehat{Q}(s,a)$ and $n(s,a)$.

---

END-OF-PROBE $(s : \text{state}, d : \text{depth})$
  **if** $s \notin \mathcal{T}$ **then** add $s$ to $\mathcal{T}$
  **if** $n(s) > 0$ **or** CONVERT$(s)$ **then return** false
  **if** $d \leq \sigma$ **then**
    $\sigma \leftarrow -1$     // dummy value, to prevent node update
    $retract \leftarrow$ true
  **return** true

---

EVALUATE $(s : \text{state}, d : \text{depth})$
  **if** $|\Pi_s^A| < \phi$ **then**
    $\pi \leftarrow$ GENERATE-POLICY$(s, H - d)$
    add $\pi$ to $\Pi_s^A$   // and thus to $\Pi_s$
  **else**
    $\pi \sim \mathcal{U}\left[\Pi_s^A\right]$
  **for** $t \leftarrow d \ldots H$ **do**
    $a \sim \pi(s,t)$
    $s' \sim P(S \mid s, a)$
    $r \leftarrow r + R(s, a, s')$
    $s \leftarrow s'$
  UPDATE-NODE$(s, \pi, r)$
  **if** $\frac{\widehat{\text{Var}}(s,\pi)}{n(s,\pi)} < \psi$ **then** remove $\pi$ from $\Pi_s^A$
  **return** $r$

---

UPDATE-NODE $(s: \text{state}, x: \text{action } or \text{ policy}, r : \text{reward})$
  $n(s,x) \leftarrow n(s,x) + 1$
  $\delta \leftarrow r - \widehat{Q}(s,x)$
  $\widehat{Q}(s,x) \leftarrow \widehat{Q}(s,x) + \frac{\delta}{n(s,x)}$
  $\widehat{\text{Var}}(s,x) \leftarrow \frac{\widehat{\text{Var}}(s,x) \cdot (n(s,x) - 2) + \delta \cdot (r - \widehat{Q}(s,x))}{n(s,x) - 1}$

---

SWITCH-FUNCTION $(n : \text{iteration}, \sigma \in [\![H]\!])$
  **if** $retract$ **or** $\sigma = H$ **then** $\sigma \leftarrow 0$
                        **else** $\sigma \leftarrow \sigma + 1$
  $retract \leftarrow$ false
  **return** $\sigma$

---

EXPLORATION-POLICY $(s : \text{state})$
  **return** $a \sim \mathcal{U}[A(s)]$

---

ESTIMATION-POLICY $(s : \text{state})$
  **return** $a \sim \mathcal{U}\left[\{a \mid \arg\max_{a \in A(s)} \widehat{Q}(s,a)\}\right]$

---

Figure 4: MCTS2e sub-routines for BRUE$_{\mathcal{IC}}$. For the CONVERT procedure called by END-OF-PROBE, see Figure 5.

Technically, candidate evaluation is performed as follows. For each $T_{IN}$ candidate $(s,h)$, the algorithm maintains a set of policies $\Pi_s$, as well as a subset of "active" policies $\Pi_s^A \subseteq \Pi_s$, size of which is bounded by a fixed parameter $\phi$. The set $\Pi_s^A$ is used by the sub-routine EVALUATE that selects a policy from $\Pi_s^A$ for evaluation uniformly at random. In case $\phi$ allows expanding the active subset $\Pi_s^A$, EVALUATE uses a newly generated policy (e.g., by selecting actions uniformly at all states that can be reached by following the actions at preceding states). The new policy is then added to $\Pi_s^A$ (and thus to $\Pi_s$).

After a policy $\pi$ is selected by EVALUATE, it is exe-

CONVERT (s : state)

1. $m \leftarrow \sum_{\pi \in \Pi_s} n(s, \pi)$
2. $\mathbf{EE} \leftarrow \sum_{\pi \in \Pi_s} \frac{n(s,\pi)}{m} \widehat{Q}(s, \pi)$
3. $\mathbf{EV} \leftarrow \sum_{\pi \in \Pi_s} \frac{n(s,\pi)}{m} \widehat{\text{Var}}(s, \pi)$
4. $\mathbf{VE} \leftarrow \sum_{\pi \in \Pi_s} \frac{n(s,\pi)}{m} (\widehat{Q}(s, \pi) - \mathbf{EE})^2$

5. if $\frac{\mathbf{EV}}{m} \geq \mathbf{VE}$ then
      return false
   else
6.    for $a \in A(s)$ do
7.       $\pi_a \leftarrow \arg\max_{\{\pi : \pi(s) = a\}} \widehat{Q}(s, \pi)$
8.       $\widehat{Q}(s, a) \leftarrow \widehat{Q}(s, \pi_a)$
9.       $n(s, a) \leftarrow n(s, \pi_a)$
      return true

Figure 5: Implementation of the decision rule of BRUE$_{\mathcal{IC}}$ for type conversion.

Table 1: Aggregated IPPC-2011 scores from Figure 6; boldfacing indicates top performance in the respective domain (and overall, in the last row).

|  | MAB | UCT | BRUE | BRUE$_{\mathcal{I}}$ | BRUE$_{\mathcal{IC}}$ |
|---|---|---|---|---|---|
| *Traffic* | **0.97** | 0.96 | 0.77 | 0.92 | 0.95 |
| *Crossing* | 0.83 | 0.84 | 0.85 | 0.83 | **0.87** |
| *Navigation* | 0.61 | 0.69 | 0.46 | 0.65 | **0.75** |
| *Game-of-Life* | 0.78 | 0.79 | 0.68 | 0.79 | **0.82** |
| *SysAdmin* | 0.86 | 0.81 | 0.73 | 0.86 | **0.87** |
| **total** | 0.81 | 0.82 | 0.70 | 0.81 | **0.85** |

cuted once starting at $(s, h)$. The resulting cumulative reward $r$ is then used to update statistics about the particular policy $\pi$, including the empirical mean $\widehat{Q}(s, \pi)$ (which is just a more convenient for us naming for the estimator $\widehat{V}_h^\pi(s)$), the empirical variance $\widehat{\text{Var}}(s, \pi)$, and the counter $n(s, \pi)$. The reward $r$ is also used later on, in the recursive rollback of PROBE (Figure 2). When the variance of $\pi$ decreases below a predefined threshold $\psi$, it is removed from the active subset $\Pi_s^A$ to be replaced by a new policy. This mechanism ensures that all policies will eventually be sampled, which is essential to guarantee convergence of our estimates of the two sources of variance.

The sub-routine END-OF-PROBE bares similarity to this of UCT, but it is extended with enforcing T$_{\text{IN}}$-candidates to remain leaves as long as the respective CONVERT attempts come out unsuccessful. In CONVERT, the statistics about policies are used to estimate the mean of the policies' variance **EV**, the variance in the policies value **VE**, the overall mean value of the policies **EE**, and $m$, the total number of samples from all the policies in $\Pi_s$. If the evaluated T$_{\text{IN}}$-candidate $(s, h)$ meats the conversion condition, CONVERT returns true, and the newborn T$_{\text{IN}}$-forecaster $(s, h)$ is no longer forced to remain a leaf. Just before that, the standard node variables $\widehat{Q}(s, a)$ and $n(s, a)$ are initialized with the information gathered around the empirically best policy $\pi_a$ that starts with the respective action $a$. Importantly, since the candidate evaluation criterion is not blocking, all candidates eventually convert to T$_{\text{IN}}$ and act as in BRUE. This provides BRUE$_{\mathcal{IC}}$ with BRUE's quality convergence rate in long term.

At last, BRUE$_{\mathcal{IC}}$ borrows its incremental tree expansion mechanism from BRUE$_{\mathcal{I}}$: When a leaf is encountered before the switching point $\sigma$, no node is updated, and a global flag *retract* is set to true, signaling SWITCH-FUNCTION to set the switching point to 0 in the subsequent iteration. This way, the switching point is selected systematically as in BRUE, but now ranging between the root and a leaf rather than on the entire horizon.

This finalizes the description of BRUE$_{\mathcal{IC}}$. To examine its relative effectiveness, we conducted a comparative evaluation under the same experimental setup as described in Section 3. We evaluated five algorithms: MAB-Uniform, UCT, BRUE, BRUE$_{\mathcal{I}}$, and BRUE$_{\mathcal{IC}}$. The results are presented in Figure 3 and summarized in Table 1; all scores in Table 1 are within confidence bounds of $\pm 0.01$. For the sake of readability, here we excluded Random from the presentation. Overall, BRUE$_{\mathcal{IC}}$ exhibited a robustly good performance across the domains, finishing top performer on most problem instances. BRUE$_{\mathcal{IC}}$ is also consistently better than BRUE$_{\mathcal{I}}$, indicating that selective type conversion plays an important role in its performance. We also notice that, in most cases, MAB-Uniform was not very far off from the leader, falling short only in few instances of *Navigation*. Since MAB-Uniform can be seen as BRUE$_{\mathcal{IC}}$ with an ultra-conservative, unsatisfiable condition for type conversion, this suggests that further introspection into the specific decision rule of BRUE$_{\mathcal{IC}}$, as well as into the impact of the parameters in its implementation, should be valuable.

Finally, we examine the simple regret reduction of BRUE$_{\mathcal{IC}}$ under varying time budgets, using the experimental settings for *Sailing* domain from [11]. In the bottom part of Figure 6, we plot the simple regret of the actions recommended by MAB-Uniform, UCT, BRUE, and BRUE$_{\mathcal{IC}}$ under different planning time windows, averaged over 300 instances of $10x10$ and $20x20$ grid sizes. The results reveal that not only does BRUE$_{\mathcal{IC}}$ significantly outperforms UCT uniformly over time, right from the beginning of deliberation when BRUE is still lagging behind, but also that its simple regret reduction rate is rather comparable to BRUE's in the longer term.

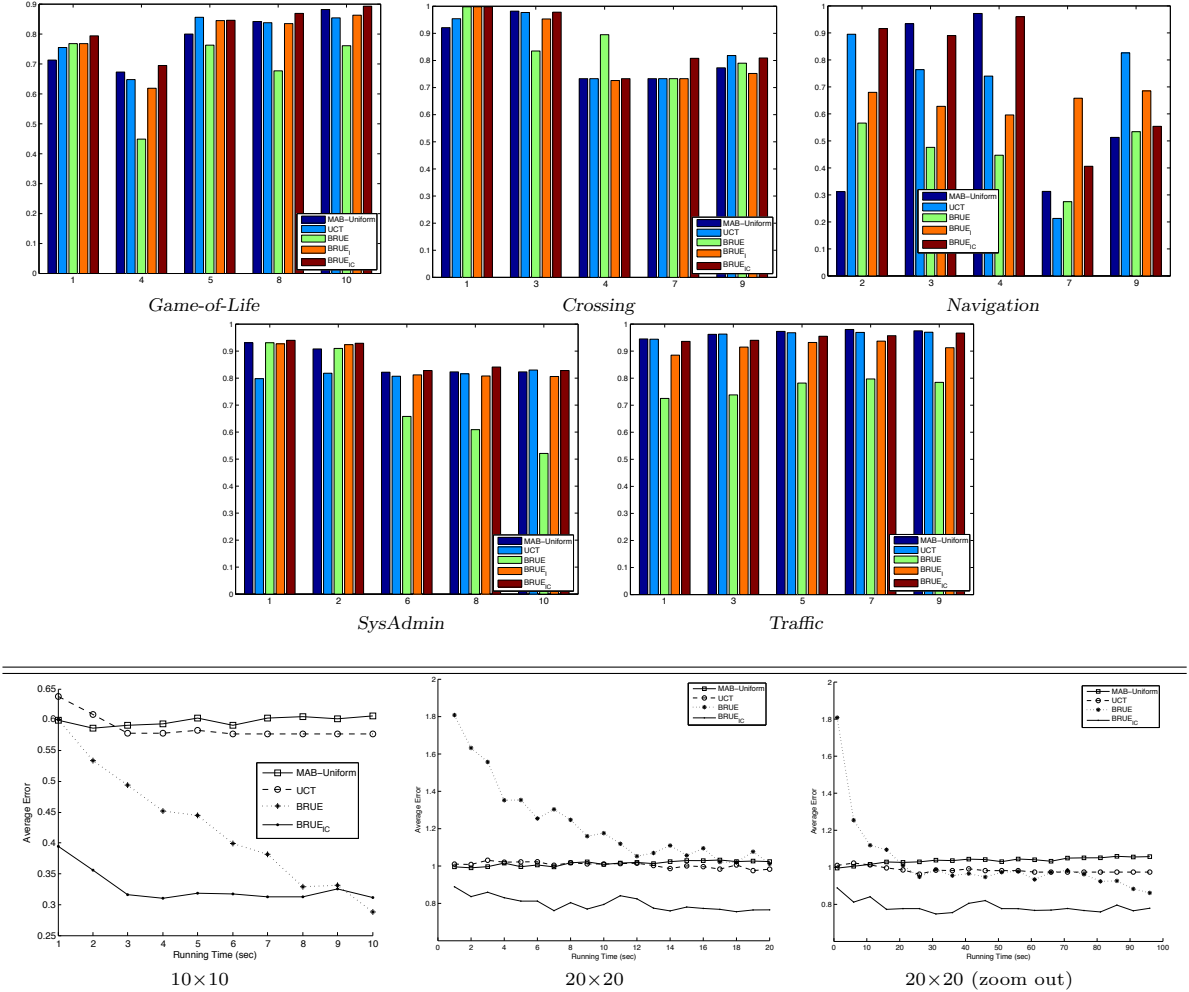

Figure 6: Performance of the MCTS algorithms with and without selective tree expansion. Top: Per-instance IPPC-2011 scores under the experimental setup as in Section 3. Bottom: Absolute performance of the algorithms in terms of average simple regret on the *Sailing* domain problems with 10×10 and 20×20 grid maps.

## 6 SUMMARY

Considering online planning for generative MDPs, we have investigated and combined the high-level principles that underly different computational schemes for this problem, and showed that their individual strengths can be put together at work. We then rationalized a principle of selective tree expansion that aims at automatically adapting Monte-Carlo exploration to the specifics of the MDP in hand, and suggested a concrete implementation of this principle within Monte-Carlo tree search methods. The resulting algorithm, $BRUE_{\mathcal{IC}}$, favorably competes with other MCTS algorithms under short planning times, while preserving the attractive convergence properties of the (not so effective under short planning windows) algorithm BRUE [11], as well as the empirical strength of the latter under permissive planning windows.

In previous works, some forms of selective tree expansion has been shown extremely effective in *forward-search* planning for declarative MDPs [7, 4]. In that context, our work can be seen as the first selective tree expansion framework for Monte-Carlo planning, which is applicable in generative MDPs as well. Likewise, our work joins some other recent techniques for enhancing Monte-Carlo planning with adaptivity to the given problem, such as hindsight optimization for declarative MDPs [17, 26], and metalevel decision procedures for sample routing for both declarative and generative MDPs [13]. At the high level, these techniques appear complementary to selective tree expansion, and thus studying their interplay and cumulative value is a promising venue for future research.

**Acknowledgements** This work was partially supported by the EOARD grant FA8655-12-1-2096, and the ISF grant 1045/12.